\documentclass{article}

\PassOptionsToPackage{numbers, compress}{natbib}

\usepackage[preprint]{tackling_climate_workshop_style}
\usepackage{xcolor}
\usepackage{subfiles}
\usepackage{rotating}
\usepackage{enumitem}
\usepackage{graphicx}
\usepackage{subcaption}
\usepackage{changepage}
\usepackage{comment}
\usepackage{wrapfig}
\usepackage[utf8]{inputenc} 
\usepackage[T1]{fontenc}    
\usepackage{hyperref}       
\usepackage{url}            
\usepackage{booktabs}       
\usepackage{amsfonts}       
\usepackage{nicefrac}       
\usepackage{microtype}      

\title{\textit{Top-down Green-ups: Satellite Sensing and Deep Models to Predict Buffelgrass Phenology}}

\author{%
  Lucas Rosenblatt\thanks{Equal contribution.} \\
  New York University\\
  \And
  Bin Han$^*$\\
  University of Washington \\
  \And
  Erin Posthumus  \\
  USA National Phenology Network \\
  \And
  Theresa Crimmins \\
  USA National Phenology Network \\
  \And
  Bill Howe \\
  University of Washington \\
}

\begin{document}

\maketitle

\begin{abstract}
An invasive species of grass known as ``buffelgrass'' contributes to severe wildfires and biodiversity loss in the Southwest United States. We tackle the problem of predicting buffelgrass ``green-ups'' (i.e. readiness for herbicidal treatment). To make our predictions, we explore temporal, visual and multi-modal models that combine satellite sensing and deep learning. We find that all of our neural-based approaches improve over conventional buffelgrass green-up models, and discuss how neural model deployment promises significant resource savings.
\end{abstract}

\section{Introduction}\label{sec:intro}
\vspace{-0.2cm}
Increasingly severe wildfire seasons dominate headlines; these out-of-control blazes can be among the most visible results of ongoing climate change \cite{nolan2022increasing, Canon_2022}. In the Southwest United States, wildfires are made worse by an invasive plant, \textit{Pennisetum ciliare}, known commonly as buffelgrass, because the grass burns twice as hot as native vegetation \cite{mcdonald2013creating}. Furthermore, buffelgrass is a threat to biodiversity in many environments \cite{marshall2012buffel} and poses a particular threat to the Sonoran Desert, where many plants are not fire adapted; the resulting grass-fire cycle excludes native plants and may result in a conversion of Sonoran Desert shrubland to monoculture grassland with increasingly severe wildfires \cite{wilder2021grassification}.

In this paper, we remotely predict buffelgrass readiness for herbicidal treatment using a combination of satellite sensing and deep learning. Wildfire risk from buffelgrass can be reduced through effective herbicidal treatments, but only after the plant ``greens-up'' from favorable conditions to at least $50\%$ greenness \cite{gerst2021usa}. Unfortunately, predicting buffelgrass green-up is a difficult task, due to varying precipitation, soil substrates, sun exposure, elevation, geographic location, etc. \cite{ap-wallace, wallace2014predicting, deAlbuquerque2019ClimateCA}. Small groups of dedicated volunteers and professionals rely on conventional precipitation-only models to allocate resources, but these models suffer from wastefully high false-positive rates \cite{gerst2021usa}. Acknowledging urgency, the U.S. Secretary of the Interior recently announced \$200,000 in additional funds for Saguaro National Park buffelgrass treatments alone \cite{grissom2022restoring}; however, estimates suggest that treating the entire park would would cost tens of millions of dollars \cite{grissom2019facts}. Our paper addresses a pressing need for intelligent green-up predictions in the fight against buffelgrass. 

\paragraph{Contributions} Our two main contributions are: (1) improvements over conventional buffelgrass green-up models using deep learning models and satellite sensing for climate and imagery and (2) an analysis of the climate factors that influence our model's predictions. We are not the first to note that remote sensing offers a way to observe buffelgrass over large tracts of land, although as far as we are aware we are the first to employ satellite imagery and climate modeling to predict \textit{phenological} changes in buffelgrass \cite{ap-wallace,zhao2021progress}. We improve on currently deployed models (based on accumulated precipitation) by significant margins thanks at least in part to new signal sources and better temporal pattern matching, and believe that analogous approaches to ours could be fruitfully deployed for detecting phenological changes in a variety of flora beyond buffelgrass. We open source our datasets and training procedures at this repository: \textcolor{blue}{\href{https://github.com/lurosenb/phenology_projects}{https://github.com/lurosenb/phenology\_projects}}

\section{Related Work}\label{sec:related-work}
\vspace{-0.2cm}
A comprehensive study by \citet{williams2019observed} found that the clearest link between anthropogenic climate change and wildfires is through increased ``warming‐driven fuel drying.'' Buffelgrass is an example of dry, organic fuel: it dehydrates for three-fourths of the year, becoming yellowish and tinder-dry despite still being alive, and most worryingly it can double in cover every 2-3 years \cite{wayne2004buffelgrass, fieldguide}. Recent work has highlighted the effectiveness of herbicidal treatment, but explicitly calls for increased detection efforts and better resource allocation \cite{li2023effectiveness}. 
Field work and heuristic models based on weather data have been shown to predict the greenness or spread of known sites with reasonable accuracy\cite{olsson2012constancy, gerst2021usa, crimmins2022science}. Leveraging publicly available remote sensing technology has also been used in the fight against buffelgrass; \citet{zhao2021progress}, 
\begin{wrapfigure}{r}{0.33\textwidth}
  \includegraphics[width=0.32\textwidth]{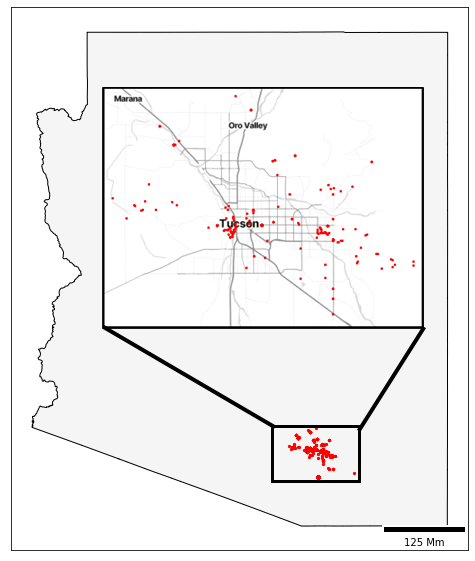}
  \caption{Clustering of buffelgrass observations in Arizona made by USA-NPN staff and volunteers.}
\end{wrapfigure}
for example, address the problem of buffelgrass \textit{site} \textit{detection} using UAVs and satellite imagery. Other prior work has also examined detecting unobserved buffelgrass sites using remote imagery, but does not explicitly model the readiness of those sites for herbicidal treatment \cite{satellite-elkind, brenner2012segmentation, marshall2014detecting}. 

Green-up prediction requires phenological observations. A primary source of observations is the USA National Phenology Network (USA-NPN), a national-scale monitoring and research initiative focused on collecting research-grade phenological data, information, and forecasts \cite{crimmins2022science}. USA-NPN offers a ``Buffelgrass Pheno Forecast,'' which currently uses the baseline accumulated precipitation model based on thresholding \cite{ap-wallace} and forecasts 50\% green up at 4km resolution across the state of Arizona \cite{gerst2021usa}. Additionally, USA-NPN facilitates the collection of high-quality observations of plant and animal life cycle stages through \textit{Nature’s Notebook}, a freely available phenology citizen science platform and dataset used by thousands of volunteer and professional scientists. Buffelgrass is one of over 1,700 species participants can observe at locations of their choosing \cite{rosemartin2014organizing}.

\section{Methodology}\label{sec:methods}
\vspace{-0.2cm}
\paragraph{Datasets}
We leverage three distinct sources of data. First, we use volunteer contributed buffelgrass phenology observations (e.g. green-up percentage $\geq \%50$) from hundreds of unique locations, gathered using USA-NPN's \textit{Nature's Notebook} platform. Second, we gather climate data from \textit{ERA5-Land}, which is a reanalysis dataset with 50 land variables spanning several decades \cite{munoz2021era5}. Of these variables, we selected 16 variables for experimentation based on domain expertise; they are listed in Table \ref{tab:external-features} in the appendix. Third,
we utilize satellite images provided by \textit{PlanetAPI}\footnote{\url{https://www.planet.com/} --- We thank Planet Labs for providing us with limited research access.}, which offers higher-resolution (3.7 meters per pixel) than standard public domain satellite imagery like Sentinel. A more detailed dataset description as well as our training practices (hyper-parameters, data splits, etc.) can be found in Sections~\ref{app:datasets} and~\ref{app:training} respectively in the appendix.


\textbf{Model 1: Accumulative Precipitation Baseline (AP)} A standard, science-backed, heuristic method for predicting buffelgrass phenology, the Accumulative Precipitation model (\textbf{AP}) simply calculates the total amount of precipitation over the past 24 days \cite{ap-wallace}. If this accumulative precipitation exceeds 1.7 inches, buffelgrass is likely to have a greenness level exceeding 50\%. This method is currently deployed by USA-NPN \cite{gerst2021usa}.

\textbf{Model 2: Long Short Term Memory (LSTM)} The \textbf{AP} model may fail to capture sequential long-term patterns or dependencies effectively, and does not incorporate other environmental variables directly \cite{ap-wallace}. We theorized that the LSTM \cite{hochreiter1997long} architecture would be well-suited for modeling the correlation between buffelgrass greenness and patterns in weather during the days leading up to observation, as LSTMs have been shown to be effective at modeling temporal patterns in climate data \cite{park2020short}. Our experiments use LSTMs with two different input sequences: \textbf{LSTM (single)} only utilizes a sequence of precipitation values as input, to be more directly comparable to the \textbf{AP} model. On the otherhand, \textbf{LSTM (multi)} takes input from a sequence of many external climate feature values in addition to precipitation values (listed in Table~\ref{tab:external-features} in Appendix).

\begin{wrapfigure}{r}{0.25\textwidth}
  \includegraphics[width=\linewidth]{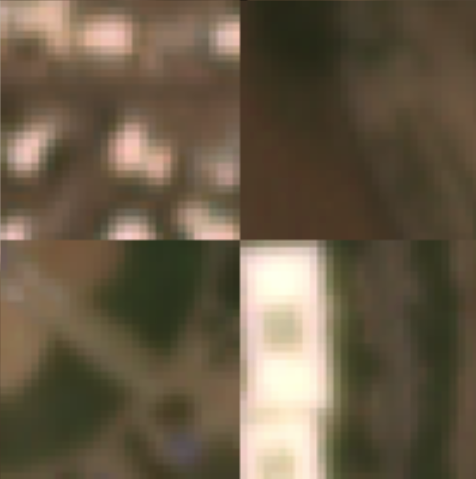}
        \caption{Four $90\times 90$ meter patches, each centered at a buffelgrass observation site.}
\end{wrapfigure}

\textbf{Model 3: Vision Transformer (ViT)} The Vision Transformer (\textbf{ViT}), as introduced in \cite{vit}, is a model primarily designed for image classification tasks. It adopts a Transformer-like architecture that operates on image patches. In our work, we fine-tuned the pre-trained model \underline{google/vit-base-patch16-224-in21k} using \textit{PlanetAPI} data.

\textbf{Model 4: Multi-Modal LSTM+ViT} The \textbf{LSTM} models are designed to process sequential features, while the \textbf{ViT} model excels at making predictions based on spatial information derived from satellite images. Recognizing the potential for these two modalities to enhance each other's performance when combined, we adopt a multi-modal approach, which has been shown to capture non-overlapping signal and improve climate models \cite{hong2020more}. Specifically, we leverage the \textbf{ViT} model to generate embeddings from satellite images. These embeddings are then concatenated with the sequential embeddings derived from external features with \textbf{LSTM}. A final linear layer then yields predictions.

\section{Results and Discussion}\label{sec:results}
\vspace{-0.2cm}
\paragraph{Neural Models Outperform Simple AP Model} The prediction results are presented in Table \ref{tab:comb-result}. Note that all of the neural models exhibit better performance compared to the simple accumulative precipitation model (AP). Explicitly modeling temporal precipitation patterns alone is enough for a significant boost, as even \textbf{LSTM (single)} demonstrates a notable enhancement in accuracy (13\% points $\uparrow$), F1 (0.26 $\uparrow$) over the AP model. With added variables, the \textbf{LSTM (multi)} achieves the best accuracy, F1, and false negative rate of all models, with smaller standard deviations, underscoring the importance of temporal features and other climate variables in modeling this task.

The \textbf{ViT} model, relying solely on satellite imagery, also improves over \textbf{AP} in accuracy (6\% points $\uparrow$), F1 (0.18 $\uparrow$), and false positive rate (7\% $\downarrow$). Despite beating \textbf{AP}, the \textbf{ViT} model underperformed the \textbf{LSTM} models. We believe this is for two reasons: (1) the observation-date-only satellite imagery lacks necessary temporal features and (2) the quality of the satellite images matters, but varies among the images obtained from the \textit{PlanetAPI}. At our current pixel resolution (1 pixel $\approx$ 3.7 meters), it is hard to detect the shapes and edges of buffelgrass (which often occurs in small clumps $< 10\times 10$ meters), spatial features that are usually helpful in image classification tasks. Furthermore, some images are distorted by artifacts of the satellite collection process, containing pure random noise (e.g., white/black pixels). Combining both \textbf{ViT} and \textbf{LSTM} yields results that surpass \textbf{ViT} alone but fall slightly short of the performance achieved by \textbf{LSTM (multi)}. Despite this, we are excited by the potential for multi-modal modeling in this domain -- see Figures~\ref{fig:venn}and~\ref{fig:overlap_images} in the appendix for an in depth discussion. With higher quality satellite imagery (such as from the Pleiades mission) and full temporal cover (i.e. images for each day leading up to observation), we believe that multi-modal models will eventually outperform \textbf{LSTM (multi)}. As it stands, this imagery is prohibitively expensive, but we hope to gain access to a higher volume and quality of satellite imagery in future project iterations.

\begin{table}[t!]
    \centering
    \tabcolsep=0.24cm
    \begin{tabular}{ |c|c|c|c|c| } 
        \hline
            \textbf{Model} & \textbf{Accuracy} & \textbf{F1} & \textbf{FP} & \textbf{FN} \\ \hline
            AP & 63.64\% (0.00\%) & 0.36 (0.00) & 26.77\% (0.00\%) & 9.60\% (0.00\%) \\ \hline
            LSTM (single) & 76.57\% (2.85\%) & 0.62 (0.06) & 17.07\% (3.29\%) & \textbf{6.36}\% (2.25\%)\\ \hline
            LSTM (multi.) & \textbf{78.08\%} (1.91\%) & \textbf{0.73} (0.02) & \textbf{6.97}\% (0.81\%) & 14.95\% (2.20\%)\\ \hline            
            ViT & 69.90\% (3.28\%) & 0.54 (0.06) & 19.09\% (3.29\%) & 11.01\% (3.94\%)\\ \hline
            LSTM (multi)  +ViT & 75.66\% (1.25\%) & 0.70 (0.02) & 8.48\% (2.00\%) & 15.86\% (1.93\%) \\ \hline            
    \end{tabular}
    \caption{Buffelgrass prediction performances. The accuracy, f1 score, false positive (FP) and false negative (FN) rates are reported. The 5-fold standard deviations are reported in the parenthesis.}
    \vspace{-0.4cm}
    \label{tab:comb-result}
\end{table}

\begin{figure}[t!]
    \centering
    \includegraphics[width=\linewidth]{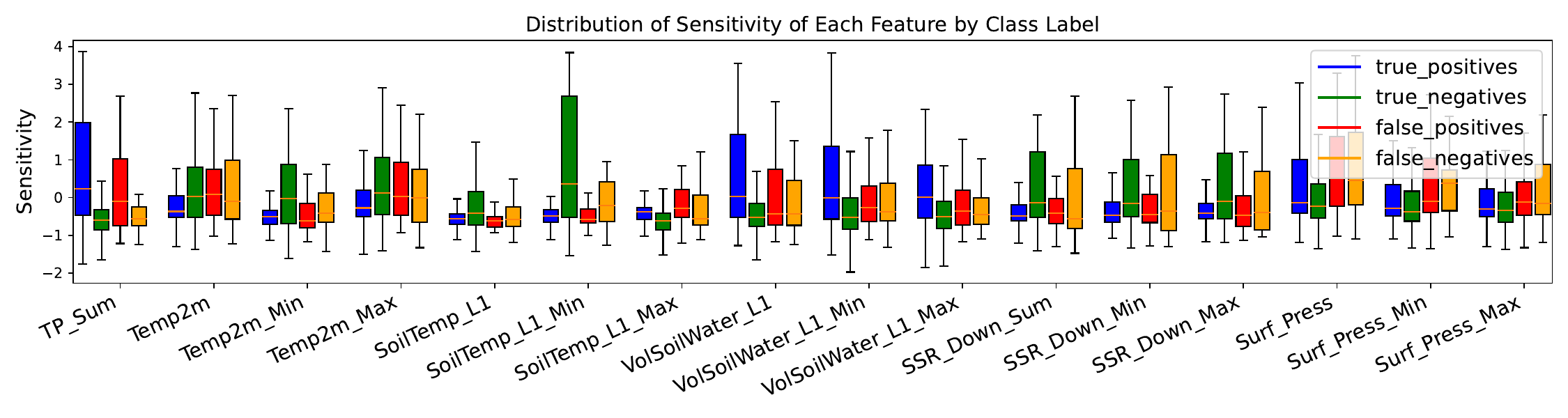}
    \caption{Aggregated sensitivities of \textbf{LSTM (multi)} across features, by confusion matrix classes. As expected, true positives are highly sensitive to \texttt{TP\_Sum, VolSoilWater\_L1} -- precipitation matters. Interestingly, true negatives are highly sensitive to \texttt{Soil\_Temp} - we conjecture that sufficiently hot days preclude green-up even with precipitation. Table~\ref{tab:feature_name_mapping} has mapping of abbreviated feature names.}
    \vspace{-0.3cm}
    \label{fig:sensitivity-horizontal}
\end{figure}

\paragraph{Feature Importance and Sensitivity Analysis}
We analyze our highest performing model, \textbf{LSTM (multi)}, with the caveat that neural networks are inherently difficult to interpret \cite{zhang2021survey}. Our primary analysis centers on model sensitivity -- procedural perturbations of training samples to elicit model behavior \cite{shu2019sensitivity}. Informally, our sensitivity analysis goes as follows: (1) for model $M$ and for training sample $x$, determine mean value $\mu_k$ for feature $k$, (2) create $x'$, which is a copy of $x$ where the value of feature $k$ in $x$ is set to $\mu_k$ (3) compute absolute difference $|M(x) - M(x')|$. We split our data into classes based on the confusion matrix to further clarify patterns \cite{shen2020designing}. In Figures~\ref{fig:sensitivity} and~\ref{fig:sensitivity2}, we show complete sensitivity results on \textbf{LSTM (multi)}. In Figure~\ref{fig:sensitivity-horizontal} we then aggregate and z-normalize results across all samples to look for global patterns; see Section~\ref{app:sensitivity} in the appendix for more analysis.

\paragraph{Resource Savings, Deployment and Impact} 
Many organizations use the USA-NPN's Buffelgrass Pheno Forecast maps to shape treatment efforts, including Pima County and the US Forest Service. These organizations manage multiple geographically diverse properties spread across southern Arizona, and green-up prediction tools shape costly crew deployments on any given day.

Models like \textbf{LSTM (multi)} can dramatically improve crew deployment accuracy, specifically because they reduce false positive predictions over \textbf{AP}. False positives are especially problematic; situations where crews are dispatched to a location but the buffelgrass is not sufficiently green for treatment wastes hundreds of dollars in crew time and vehicle fuel \cite{grissom2019facts}. False negatives are increased slightly by the neural based models, but that is more manageable because the window for herbicidal treatment (to prevent ripe seed production) is a few weeks. Model reruns with updated climate or imagery data would then still hope to turn false negatives into true positives as the phenocycle matures.

\section{Conclusion}\label{sec:conclusion}
\vspace{-0.2cm}
In this work, we tackle the problem of predicting the greenness of buffelgrass, aiming to reduce wildfire risk and protect biodiversity. We construct deep neural models that can leverage temporal information from climate features and spatial information from satellite images. Our experimental results show that: (1) neural models improve significantly over the currently deployed heuristic prediction models and (2) LSTMs with temporal climate features had the strongest predictive power of the models we tested. As satellite imagery coverage and resolution improves, one day we hope to deploy indirect climate based models alongside direct observation of greenness. For now, climate based LSTM models offer strong performance and smooth daily implementation; they are a new standard in buffelgrass green-up prediction models.

\bibliographystyle{plainnat}
\bibliography{TCCML_NeurIPS_2023_Style_File/tackling_climate_workshop}
\newpage
\section{Appendix}\label{sec:appendix}
\subsection{External Variables}

\begin{table}[hb]
    \centering
    \small
    \renewcommand{\arraystretch}{1.3}
    \begin{tabular}{|c|p{8cm}|}
        \hline
            \textbf{Variable} & \textbf{Definition} \\ \hline

            total\_precipitation\_sum & Accumulated liquid and frozen water, including rain and snow, that falls to the Earth's surface. \\ \hline

            temperature\_2m & Temperature of air at 2m above the surface of land, sea or in-land waters. \\ \hline             
            temperature\_2m\_min & daily minimum temperature\_2m value\\ \hline             
            temperature\_2m\_max & daily maximum temperature\_2m value\\ \hline

            soil\_temperature\_level\_1 & Temperature of the soil in layer 1 (0 - 7 cm) of the ECMWF Integrated Forecasting System. \\ \hline 
            soil\_temperature\_level\_1\_min & daily minimum soil\_temperature\_level\_1 value\\ \hline soil\_temperature\_level\_1\_max & daily maximum soil\_temperature\_level\_1 value\\ \hline
            
            volumetric\_soil\_water\_layer\_1 & Volume of water in soil layer 1 (0 - 7 cm) of the ECMWF Integrated Forecasting System. \\ \hline 
            volumetric\_soil\_water\_layer\_1\_min & daily minimum volumetric\_soil\_water\_layer\_1 value\\ \hline 
            volumetric\_soil\_water\_layer\_1\_max & daily maximum volumetric\_soil\_water\_layer\_1 value\\ \hline

            surface\_solar\_radiation\_downwards\_sum & Amount of solar radiation (also known as shortwave radiation) reaching the surface of the Earth. \\ \hline 
            surface\_solar\_radiation\_downwards\_min & daily minimum surface\_solar\_radiation\_downwards\_sum value\\ \hline 
            surface\_solar\_radiation\_downwards\_max & daily maximum surface\_solar\_radiation\_downwards\_sum value\\ \hline

            surface\_pressure & Pressure (force per unit area) of the atmosphere on the surface of land, sea and in-land water. \\ \hline
            surface\_pressure\_min & daily minimum surface\_pressure value\\ \hline surface\_pressure\_max & daily maximum surface\_pressure value\\ \hline
    \end{tabular}
    \caption{External features for buffelgrass greenness prediction.}
    \label{tab:external-features}
\end{table}

\begin{table}[h]
    \centering
    \renewcommand{\arraystretch}{1.2}
    \begin{tabular}{|l|l|}
        \hline
        \textbf{Original Feature Name} & \textbf{Abbreviated Name} \\
        \hline
        total\_precipitation\_sum & TP\_Sum \\
        temperature\_2m & Temp2m \\
        temperature\_2m\_min & Temp2m\_Min \\
        temperature\_2m\_max & Temp2m\_Max \\
        soil\_temperature\_level\_1 & SoilTemp\_L1 \\
        soil\_temperature\_level\_1\_min & SoilTemp\_L1\_Min \\
        soil\_temperature\_level\_1\_max & SoilTemp\_L1\_Max \\
        volumetric\_soil\_water\_layer\_1 & VolSoilWater\_L1 \\
        volumetric\_soil\_water\_layer\_1\_min & VolSoilWater\_L1\_Min \\
        volumetric\_soil\_water\_layer\_1\_max & VolSoilWater\_L1\_Max \\
        surface\_solar\_radiation\_downwards\_sum & SSR\_Down\_Sum \\
        surface\_solar\_radiation\_downwards\_min & SSR\_Down\_Min \\
        surface\_solar\_radiation\_downwards\_max & SSR\_Down\_Max \\
        surface\_pressure & Surf\_Press \\
        surface\_pressure\_min & Surf\_Press\_Min \\
        surface\_pressure\_max & Surf\_Press\_Max \\
        \hline
    \end{tabular}
    \caption{Mapping of original feature names to abbreviated names.}
    \label{tab:feature_name_mapping}
\end{table}

\subsection{Datasets}\label{dataset}
\label{app:datasets}
\vspace{-0.2cm}
\textbf{Prediction Data}: Our phenological prediction data actually comprises data from two similar sources:
\begin{itemize}[leftmargin=0.5cm]
    \item \textbf{Nature's Notebook Observation Data}: Volunteer participants have contributed buffelgrass phenology observations at hundreds of unique locations using the USA-NPN's Nature's Notebook platform. Nature's Notebook is designed to encourage repeated observations on the same individual plants over the growing season, ideally capturing green-up, flowering, seed ripening, and senescence. Individual observations consist of observation location (longitude and latitude), elevation, phenological status (whether the plant is green or not; whether the plant is flowering or not; whether fruits are present; etc.) as well as the intensity or abundance of present phenophases (if the plant is green, how much of the plant is green?) on a single date. The nature's notebook observations include 1,698 distinct $location\times date$ observations in total. According to standard practices described in \citet{gerst2021usa}, we reclassified greenness reports observations as: 1 if greenness exceeds 50\%; 0 otherwise.
    
    \item \textbf{Local One-Time Observation Data}: Because buffelgrass is highly invasive, the plants are often removed when observed, rendering repeated phenological observations within the \textit{Nature's Notebook} platform impossible. Accordingly, in 2019, the USA-NPN created an online form through which professionals and volunteers treating buffelgrass could report the instantaneous phenological status of buffelgrass plants. These ``one-time'' observations are primarily used by the USA-NPN to validate existing Pheno Forecast maps. The format of this data matches \textit{Nature's Notebook} observations, and we follow the same greenness reclassifying procedure, providing us with an additional 194 observations.
\end{itemize}

\textbf{Google Earth --- ERA5-Land Data}: To improve predictions, we use external climate or weather features that may correlate with buffelgrass greenness. ERA5-Land is a reanalysis dataset that furnishes data on 50 land variables spanning several decades at a higher resolution than the original ERA5 dataset \cite{munoz2021era5}. For more comprehensive information regarding this dataset, readers are directed to \cite{era5}. For each entry in the \textit{Nature's Notebook} dataset, we obtain external feature values from a period starting at the observation date and extending back 24 days. We have selected 16 variables for experimentation, and these are listed in Table \ref{tab:external-features}.

\textbf{Planet Satellite Images}: As discussed, several studies \cite{satellite-papp, satellite-elkind} have previously employed satellite imagery to \textit{detect} the presence of buffelgrass. In contrast, our research focuses on a different task, which is to \textit{predict} the greenness of buffelgrass using high-resolution satellite images. To facilitate this endeavor, we utilize high-resolution satellite images provided by Planet \footnote{\url{https://www.planet.com/} --- we received limited research access from the company.}, which offer a resolution of 3.7 meters. For each data record, we extract satellite images centered around the observation location point, corresponding to the observation date. Each image has the dimension of $369\times369$.  

\subsection{Training}\label{app:training}
\vspace{-0.2cm}
By combining all our prediction data sources, we were left with 987 observations to experiment with (some observations did not have accompanying satellite imagery from \textit{PlanetAPI}, as they predate the launch of the mission in 2016). We initially divided the 987 observations into a training set comprising 798 observations and a test set consisting of 198 observations. We then conducted a 5-fold cross-validation procedure on the training set to develop our models. To gauge the variability in model performance, we calculated the standard deviation across the 5 folds. 

For the \textbf{AP} model, there is no training needed. For the \textbf{LSTM} models, they are composed of two LSTM layers with a hidden dimension of 128, and a linear layer for prediction. During training, the batch size is set to 4, the learning rate to 1e-4, and the number of training epochs to 500. The \textbf{ViT} model is trained with a batch size of 2, a learning rate of $1e-5$, and it undergoes 10 training epochs. The \textbf{LSTM+ViT} model is trained with a batch size of 4, a learning rate of $1e-5$ for 100 training epochs. In all cases, a consistent random seed of 816 is employed for model training to ensure reproducibility and consistency in results. 

\section{Sensitivity Analysis}\label{app:sensitivity}
Below we provide sensitivty analysis figures for \textbf{LSTM (multi)} across every feature. Every point in each plot represents a sample in our training data, and is color coded by the confusion matrix class of the prediction that sample received by the \textbf{LSTM (multi)} model. You may immediately note that the \textbf{LSTM (multi)} model does not produce calibrated predictions i.e. the predictions are not true probabilities $\in [0,1]$, though we do threshold by 0.5. In our case, we found that taking extra steps to calibrate the model reduced model performance, which is common for deep learning models and complex architectures \cite{guo2017calibration}. Thus, it is better to think of the model outputs as likelihood scores.

Here we take a close look at each plot in order, to ascertain feature specific patterns. Note that we will delineate between sensitivity observations and conjectures about root causes.

\begin{figure}[hb]
    \centering
    \begin{subfigure}{.45\textwidth}
        \centering
        \includegraphics[width=\linewidth]{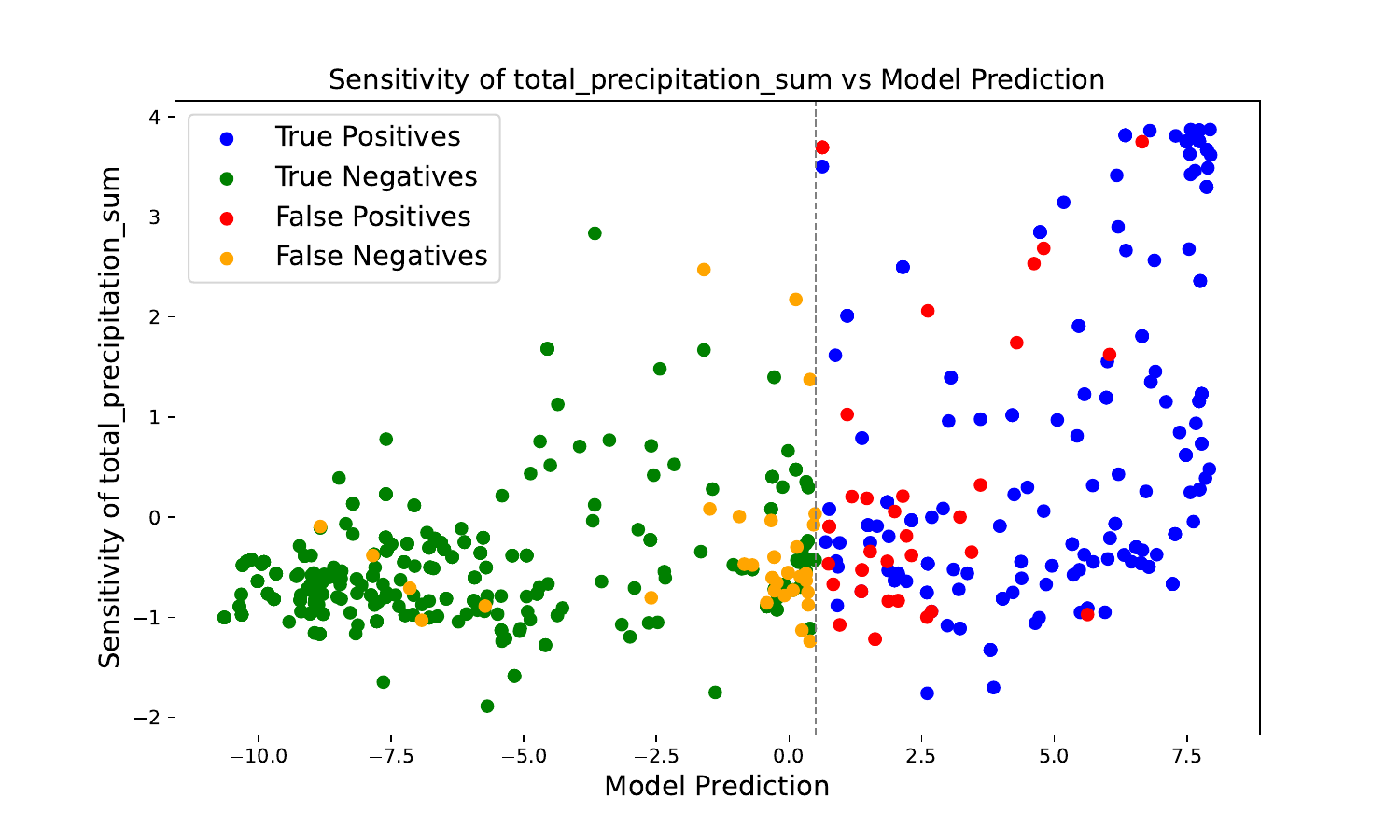}
        \caption{Total Precipitation Sum}
        \label{fig:total_precipitation_sum}
    \end{subfigure}
    \begin{subfigure}{.45\textwidth}
        \centering
        \includegraphics[width=\linewidth]{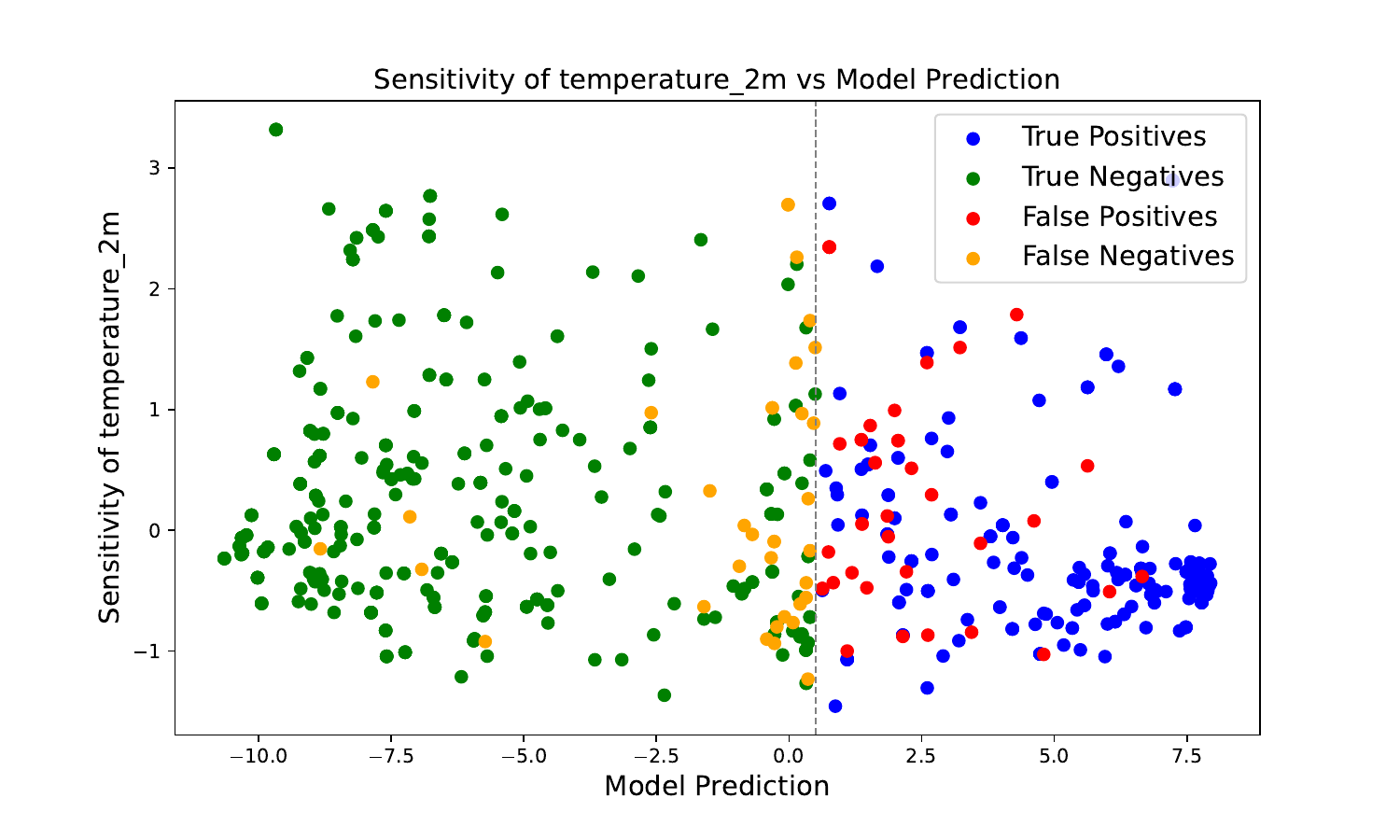}
        \caption{Temperature (2m)}
        \label{fig:temperature_2m}
    \end{subfigure}
    \begin{subfigure}{.45\textwidth}
        \centering
        \includegraphics[width=\linewidth]{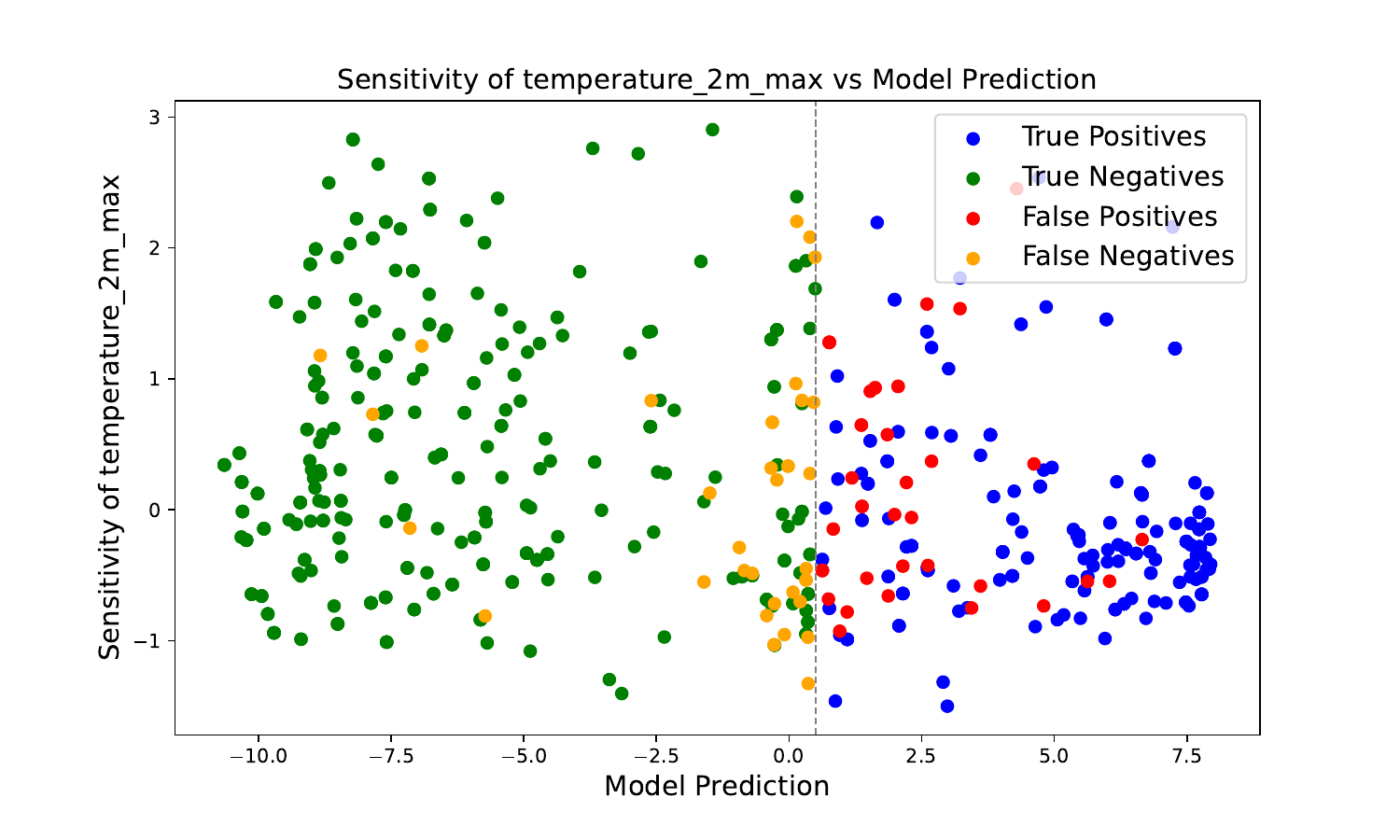}
        \caption{Temperature (2m) Max}
        \label{fig:temperature_2m_max}
    \end{subfigure}
    \begin{subfigure}{.45\textwidth}
        \centering
        \includegraphics[width=\linewidth]{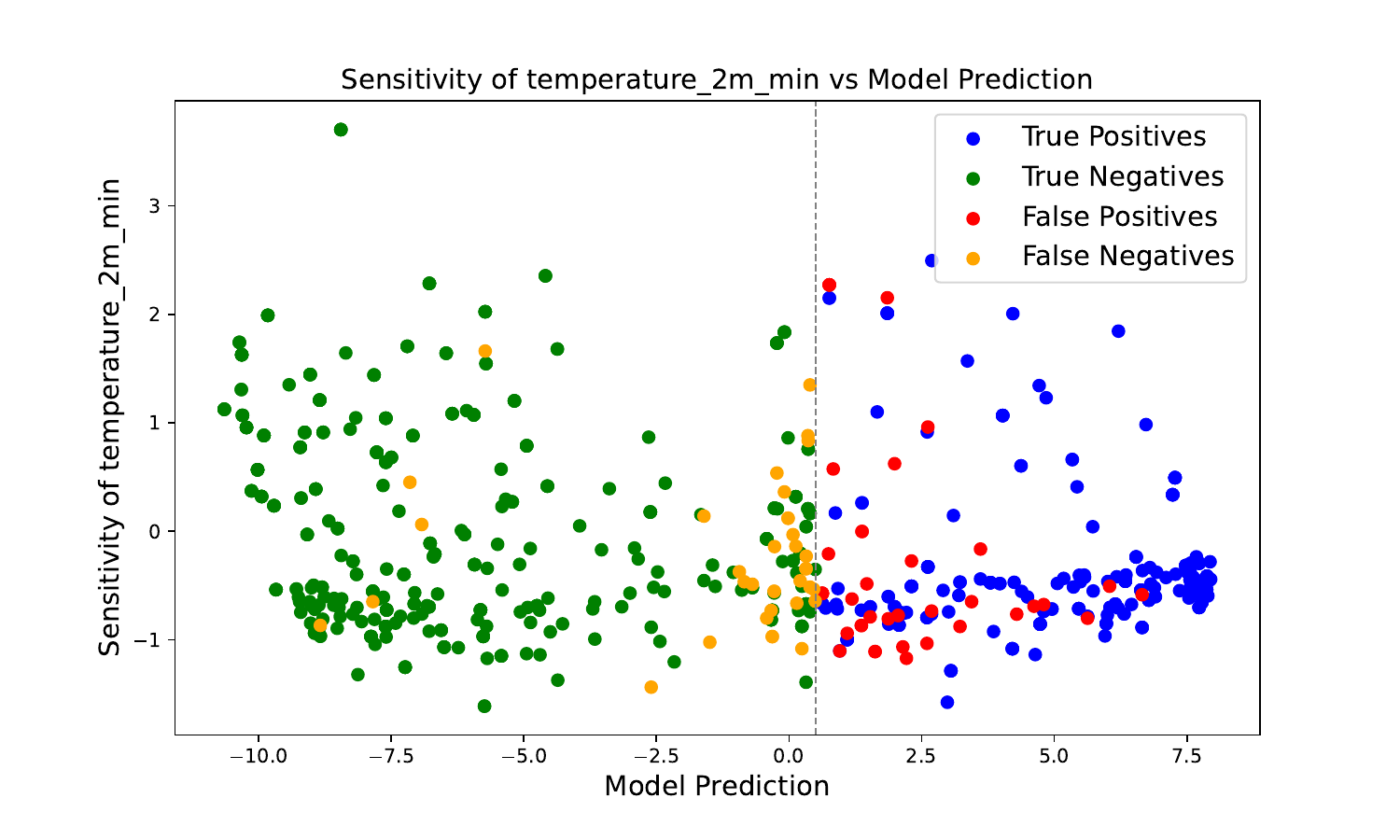}
        \caption{Temperature (2m) Min}
        \label{fig:temperature_2m_min}
    \end{subfigure}
    \begin{subfigure}{.45\textwidth}
        \centering
        \includegraphics[width=\linewidth]{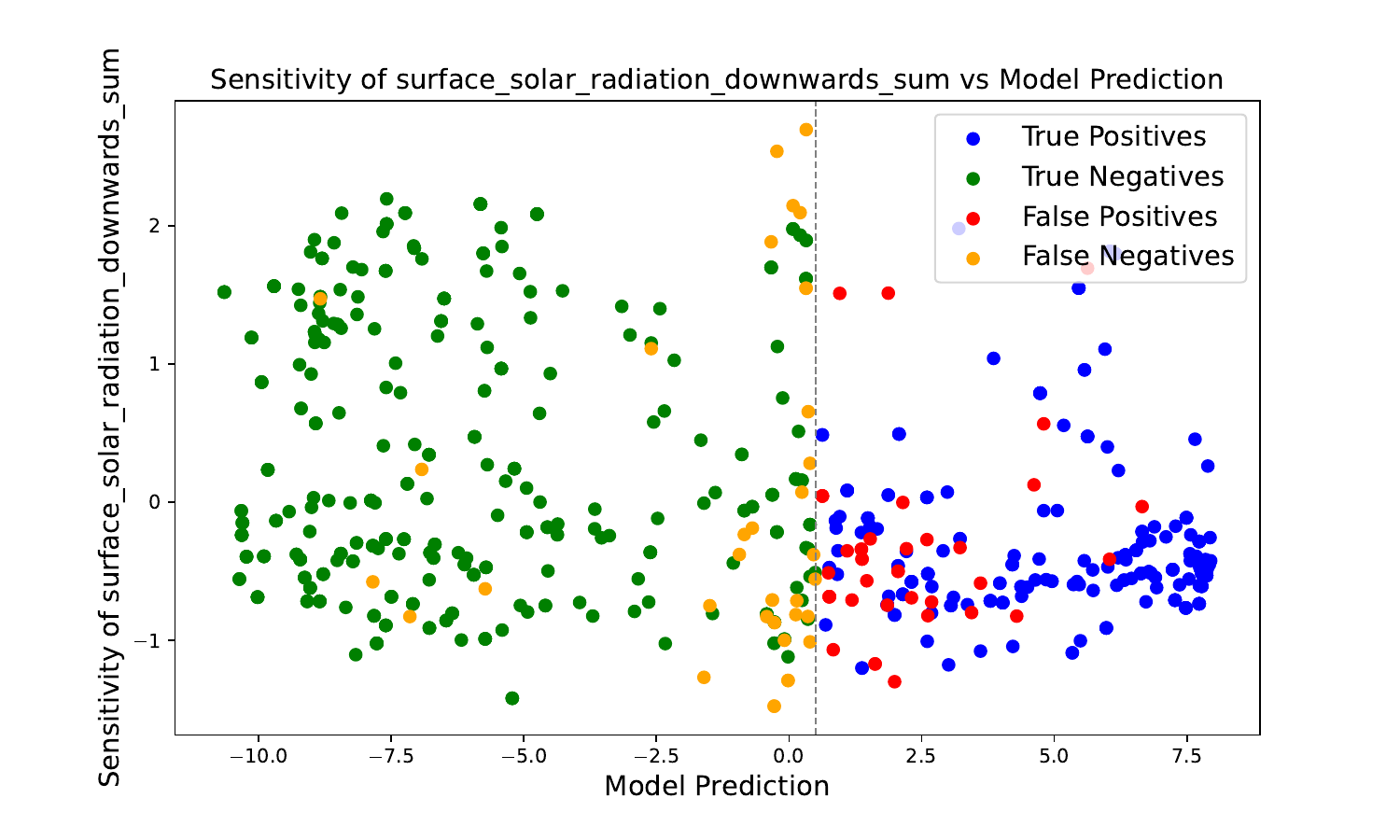}
        \caption{Solar Radiation Downward Sum}
        \label{fig:solar_radiation_downward_sum}
    \end{subfigure}
    \begin{subfigure}{.45\textwidth}
        \centering
        \includegraphics[width=\linewidth]{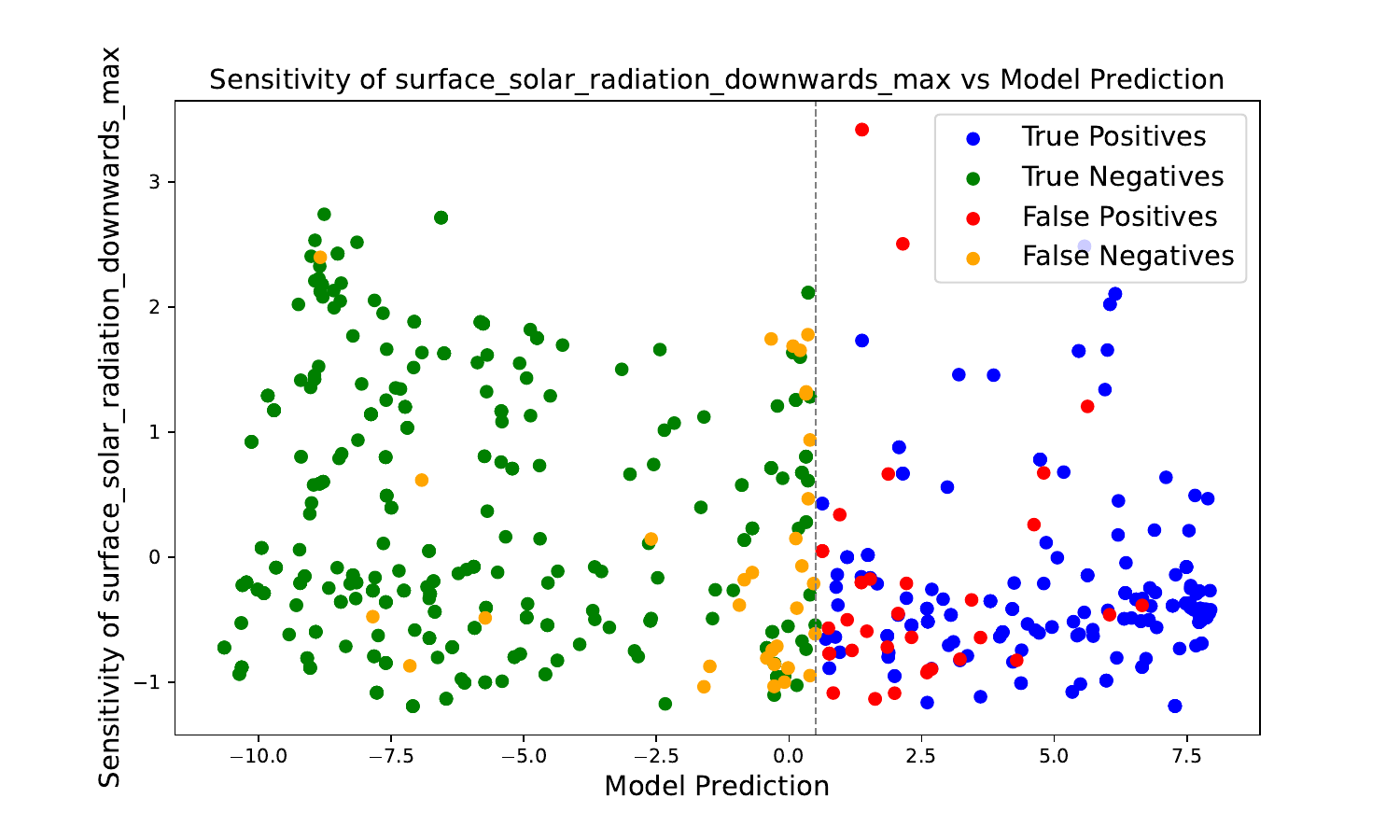}
        \caption{Solar Radiation Downward Max}
        \label{fig:solar_radiation_downward_max}
    \end{subfigure}
    \begin{subfigure}{.45\textwidth}
        \centering
        \includegraphics[width=\linewidth]{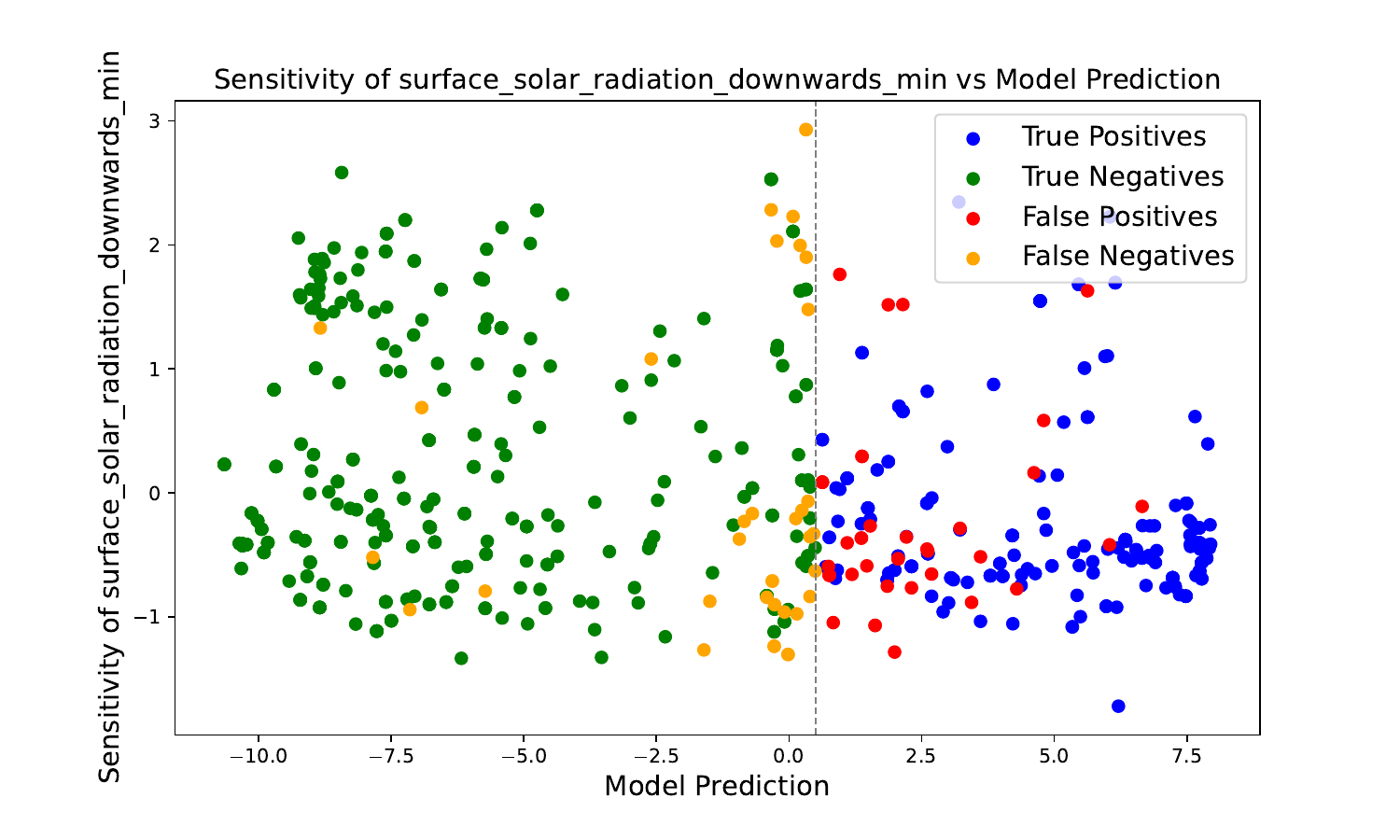}
        \caption{Solar Radiation Downward Min}
        \label{fig:solar_radiation_downward_min}
    \end{subfigure}
    \caption{Sensitivity Plots}
    \label{fig:sensitivity}
\end{figure}

\begin{figure}[hb]
    \centering
    \caption{Sensitivity Plots}
    \label{fig:sensitivity2}
    \begin{subfigure}{.45\textwidth}
        \centering
        \includegraphics[width=\linewidth]{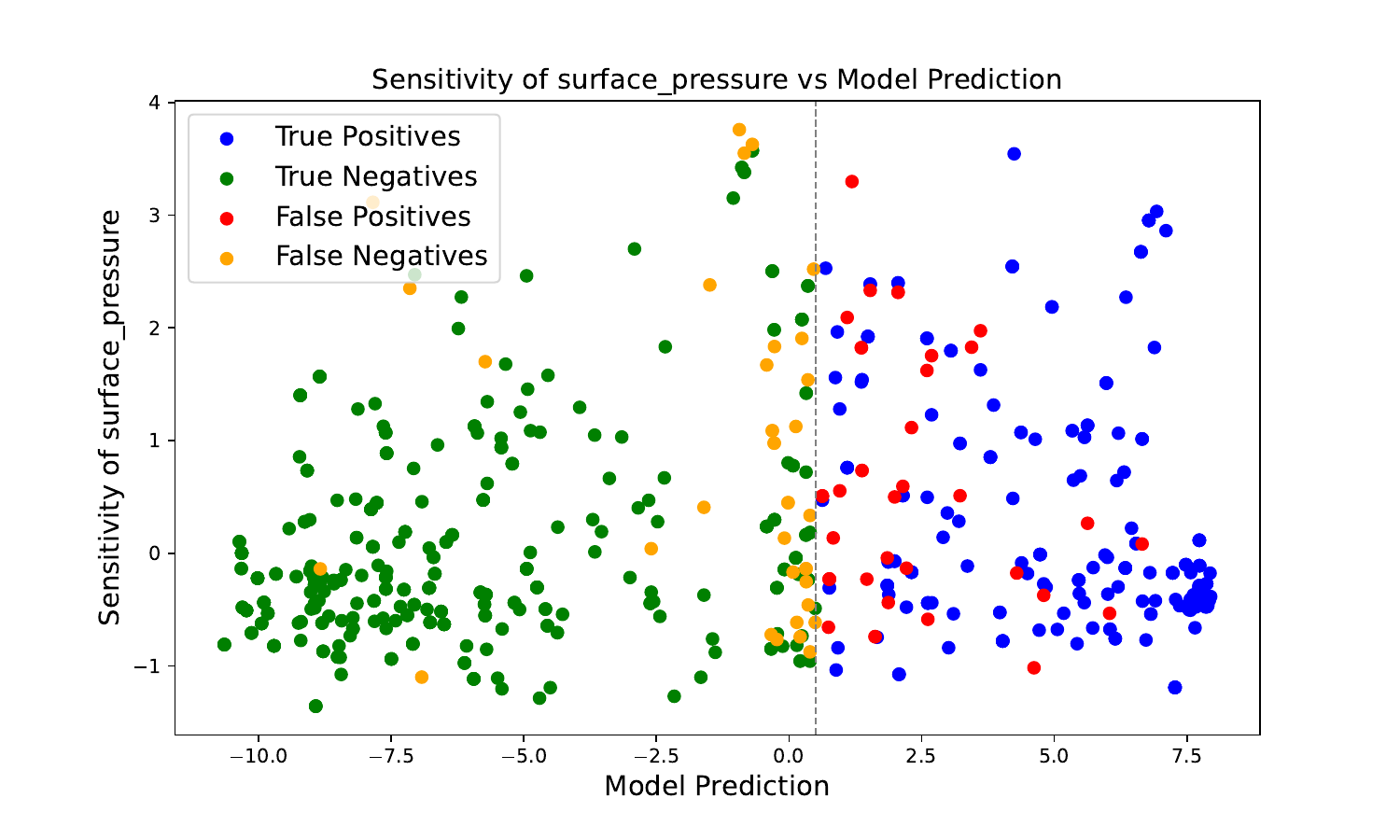}
        \caption{Surface Pressure}
        \label{fig:surface_pressure}
    \end{subfigure}
    \begin{subfigure}{.45\textwidth}
        \centering
        \includegraphics[width=\linewidth]{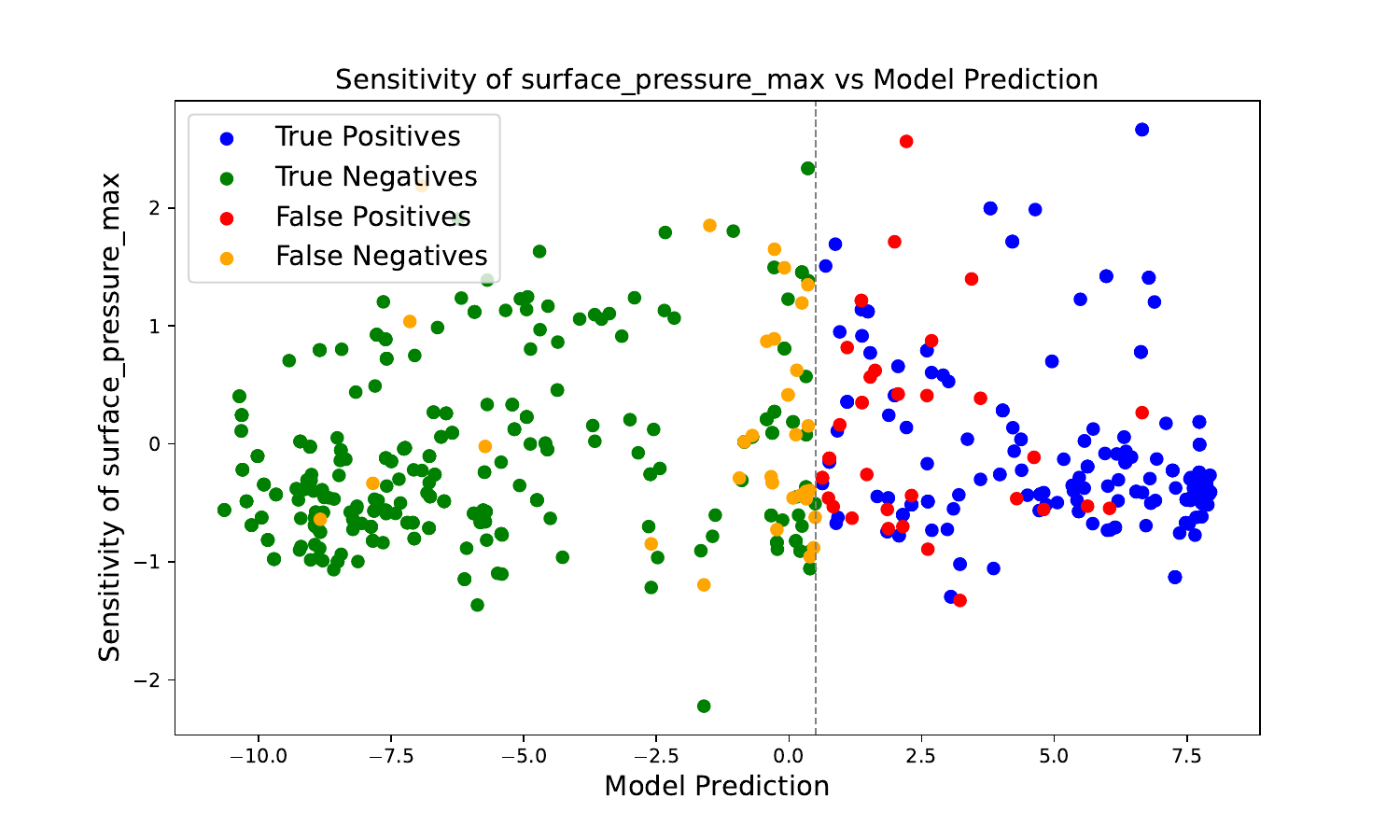}
        \caption{Surface Pressure Max}
        \label{fig:surface_pressure_max}
    \end{subfigure}
    \begin{subfigure}{.45\textwidth}
        \centering
        \includegraphics[width=\linewidth]{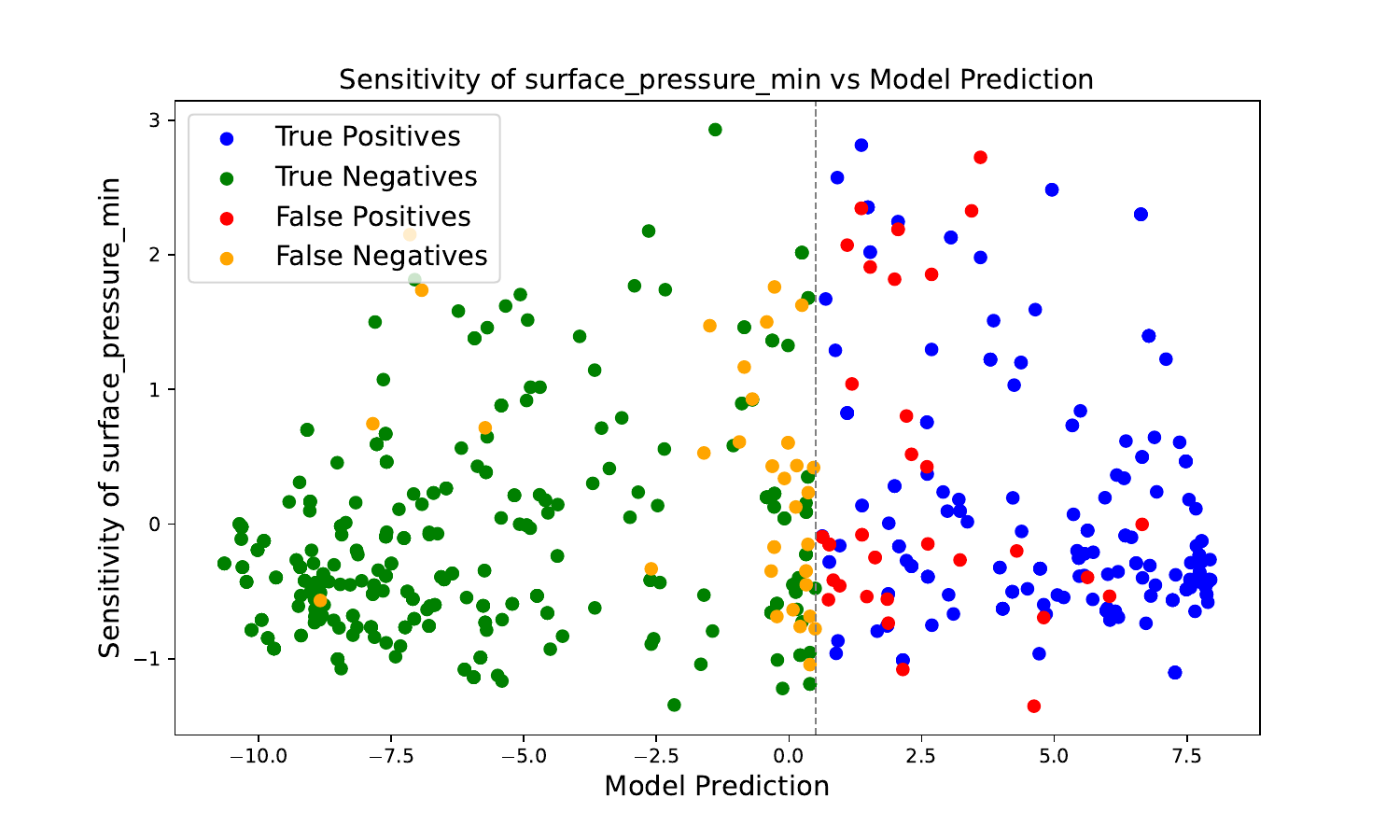}
        \caption{Surface Pressure Min}
        \label{fig:surface_pressure_min}
    \end{subfigure}
    \begin{subfigure}{.45\textwidth}
        \centering
        \includegraphics[width=\linewidth]{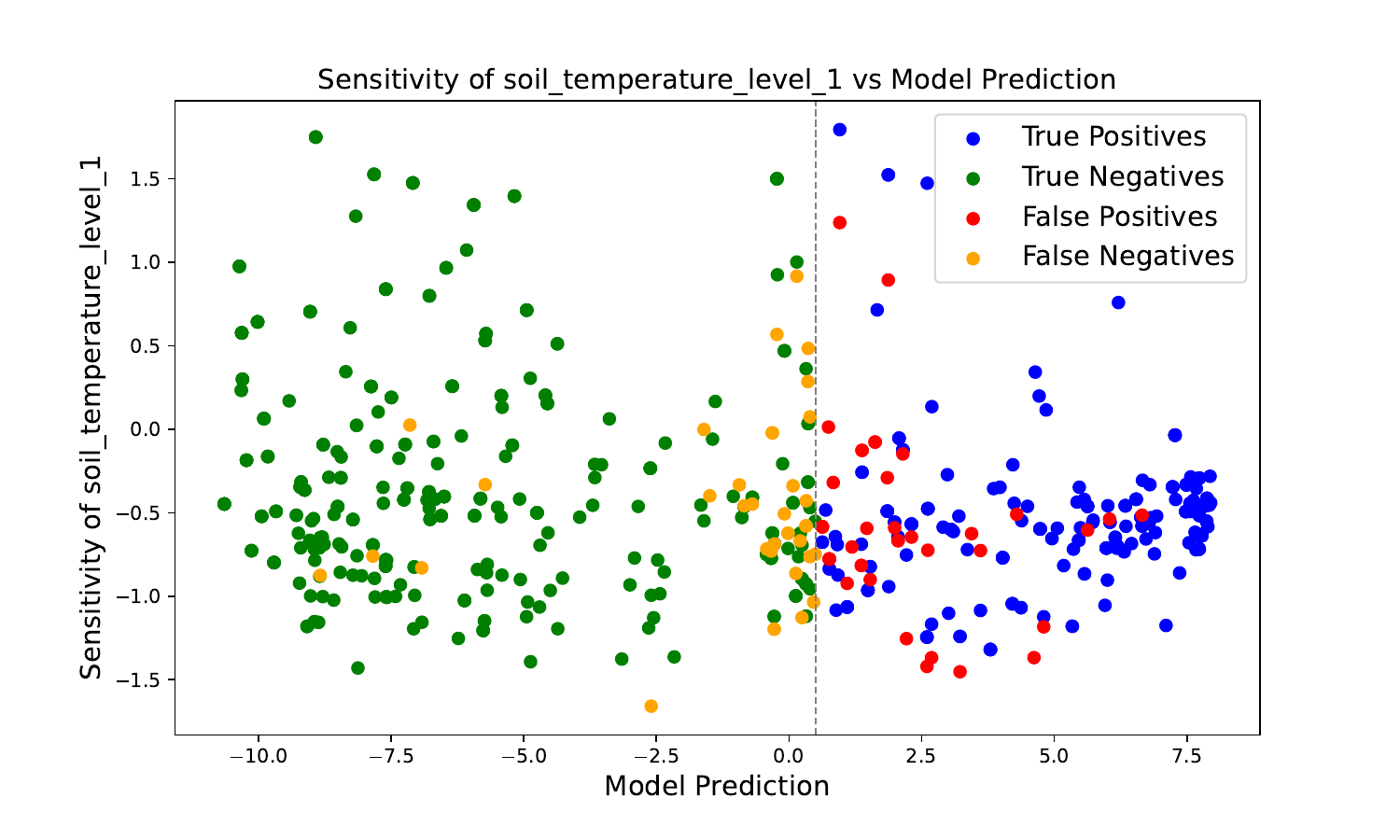}
        \caption{Soil Temperature Level 1}
        \label{fig:soil_temperature_level_1}
    \end{subfigure}
    \begin{subfigure}{.45\textwidth}
        \centering
        \includegraphics[width=\linewidth]{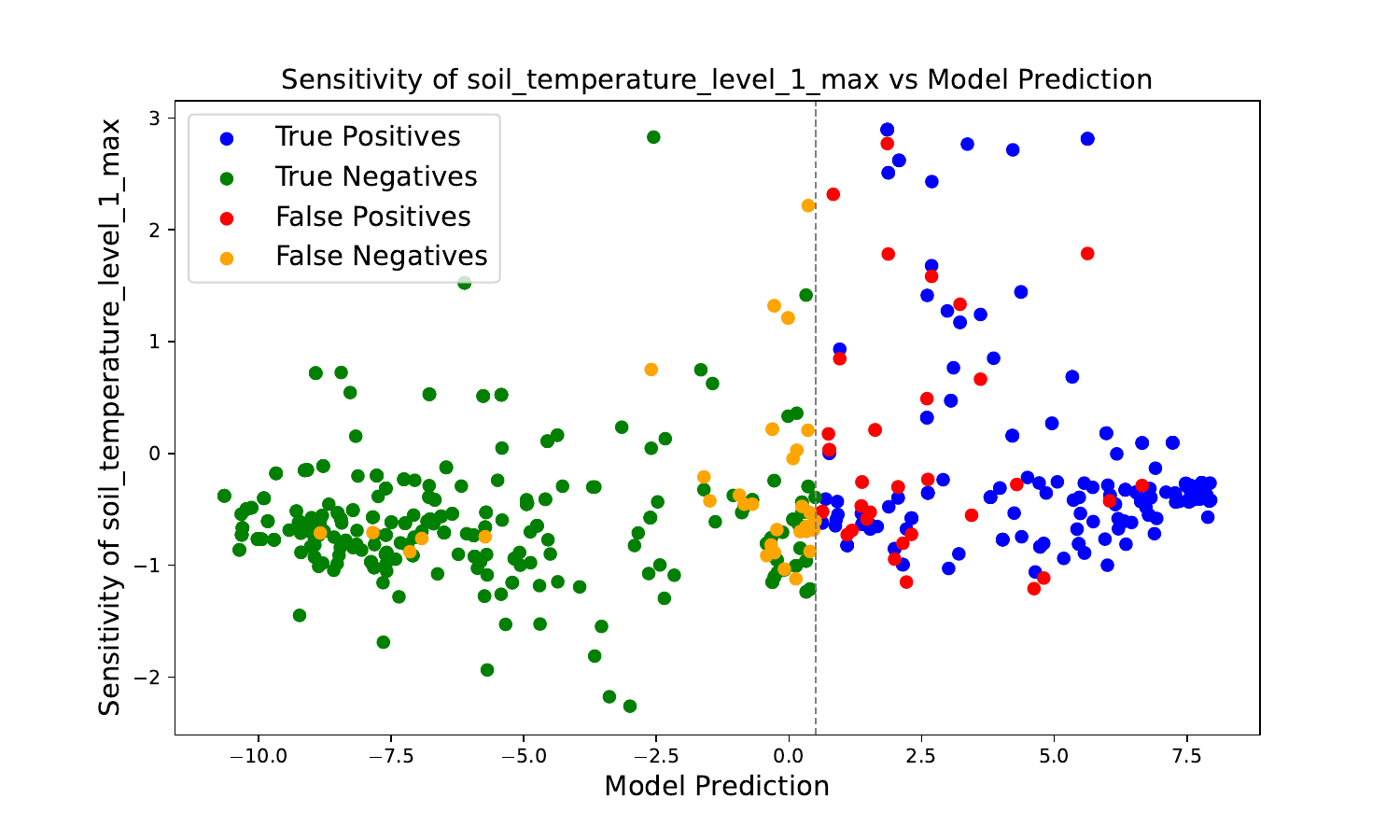}
        \caption{Soil Temperature Level 1 Max}
        \label{fig:soil_temperature_level_1_max}
    \end{subfigure}
    \begin{subfigure}{.45\textwidth}
        \centering
        \includegraphics[width=\linewidth]{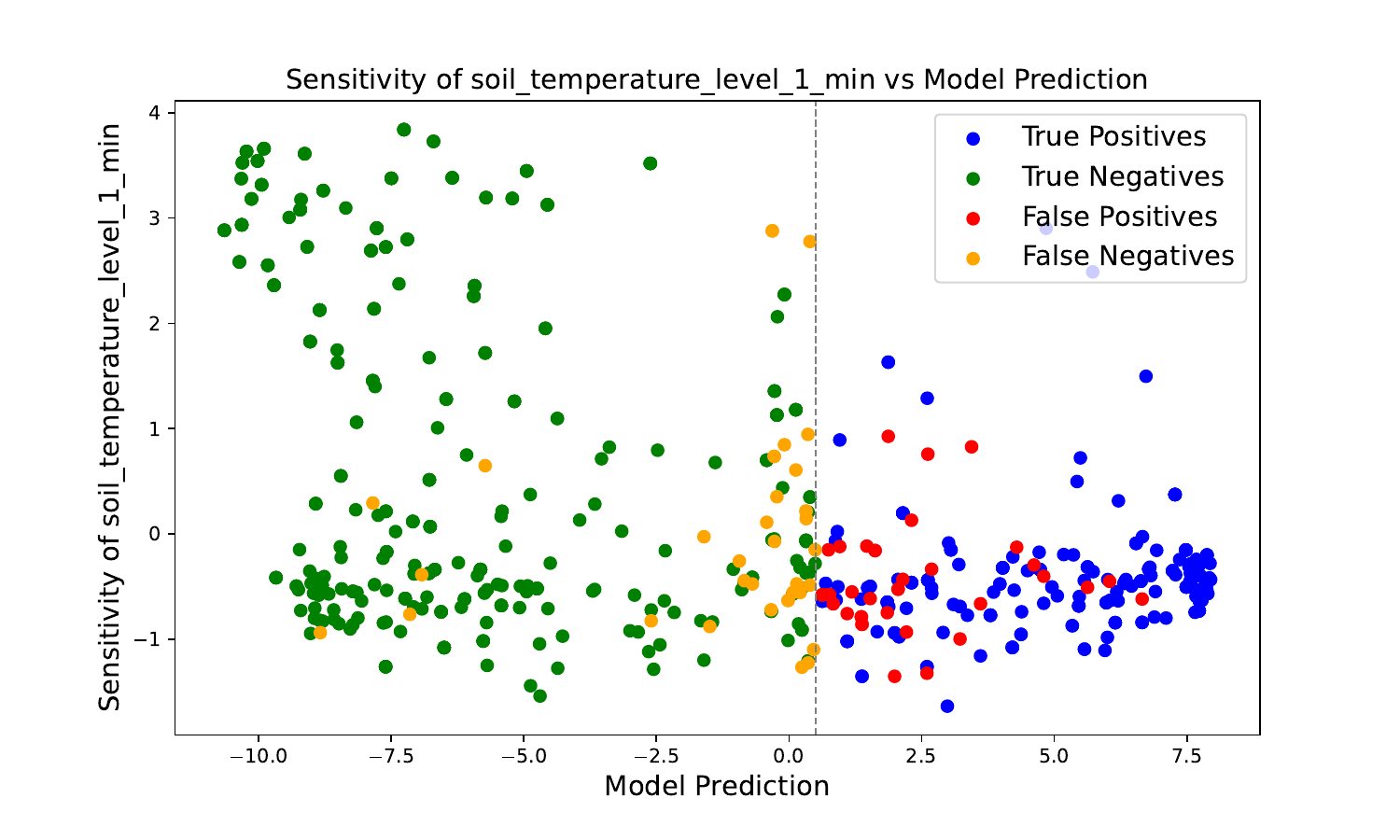}
        \caption{Soil Temperature Level 1 Min}
        \label{fig:soil_temperature_level_1_min}
    \end{subfigure}
    \begin{subfigure}{.45\textwidth}
        \centering
        \includegraphics[width=\linewidth]{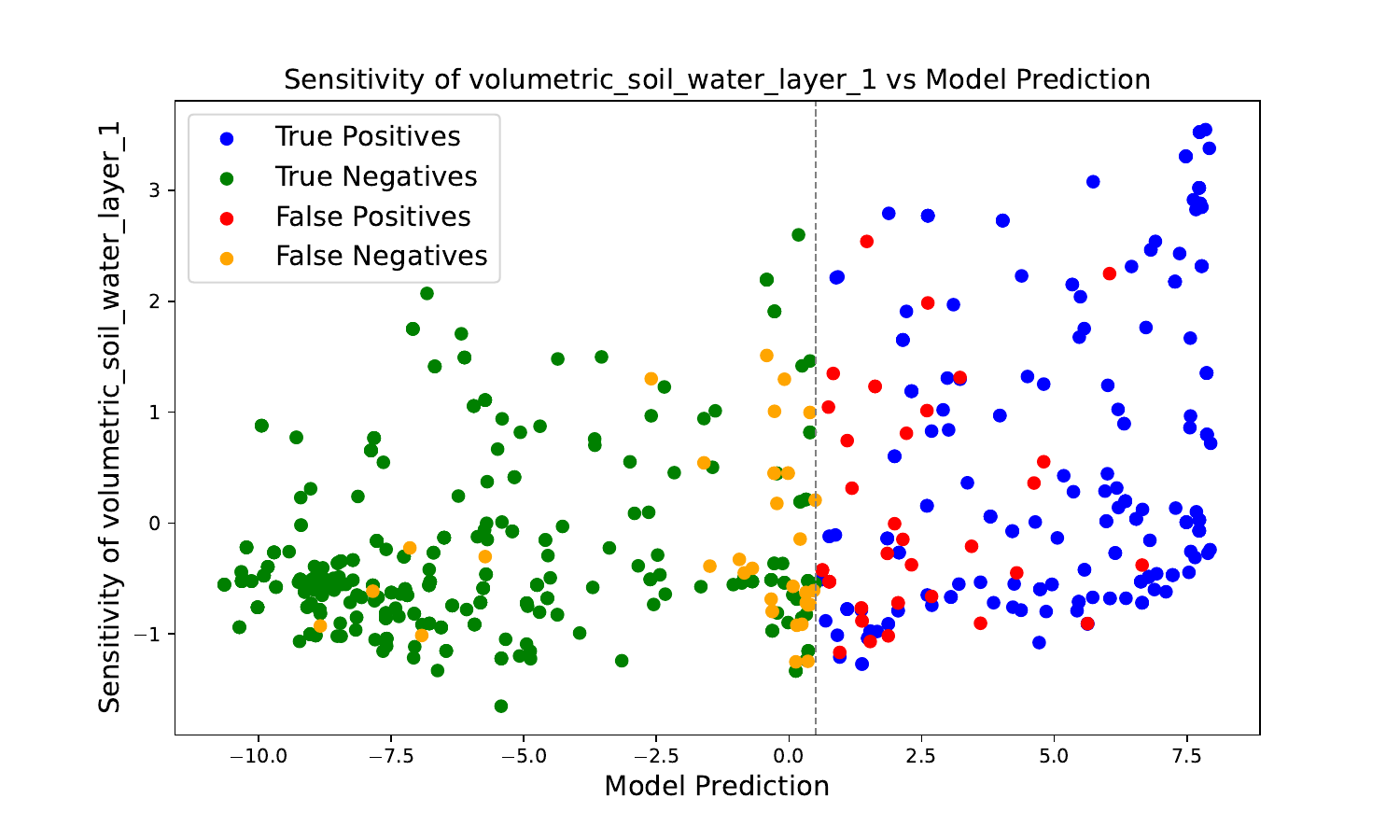}
        \caption{Volumetric Soil Water Layer 1}
        \label{fig:volumetric_soil_water_layer_1}
    \end{subfigure}
    \begin{subfigure}{.45\textwidth}
        \centering
        \includegraphics[width=\linewidth]{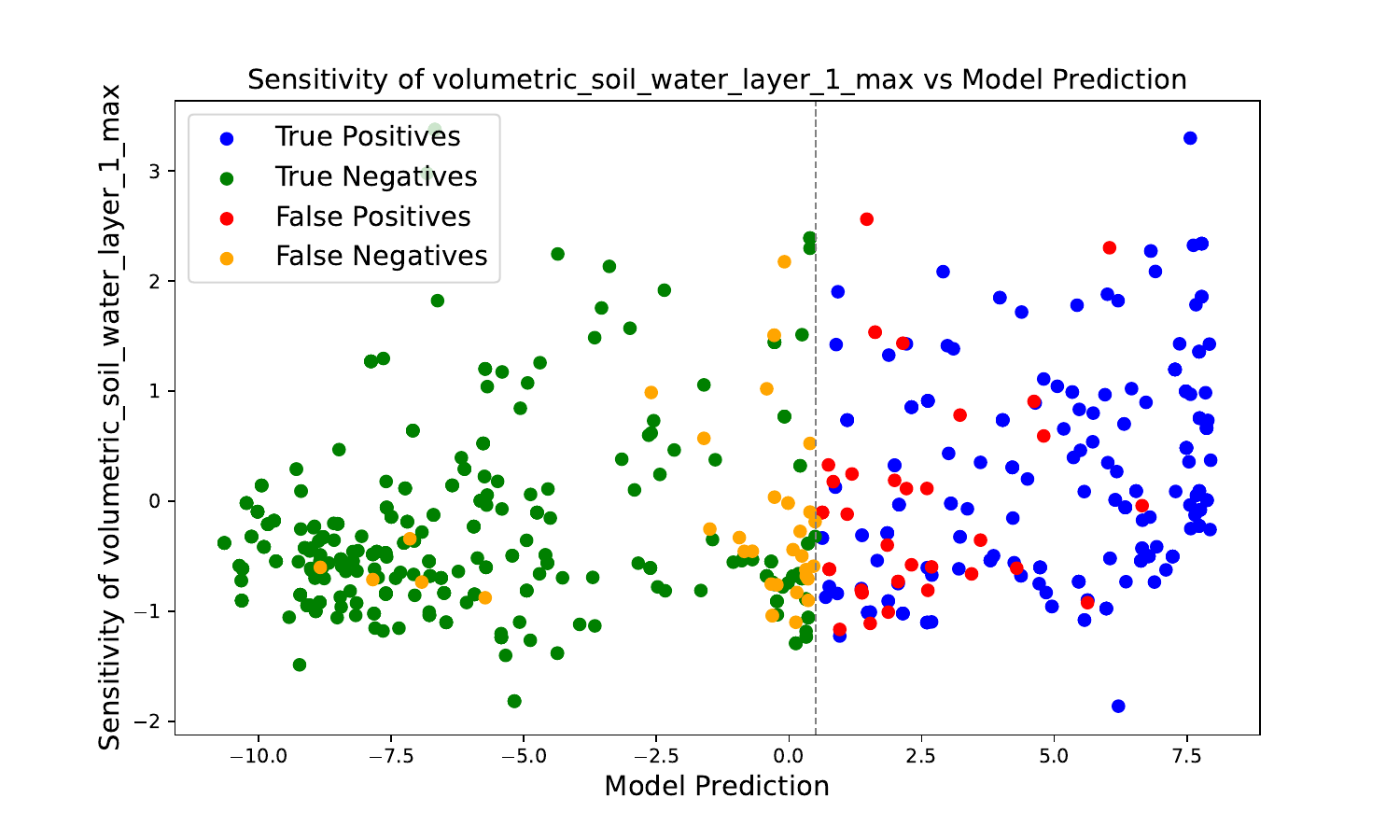}
        \caption{Volumetric Soil Water Layer 1 Max}
        \label{fig:volumetric_soil_water_layer_1_max}
    \end{subfigure}
    \begin{subfigure}{.45\textwidth}
        \centering
        \includegraphics[width=\linewidth]{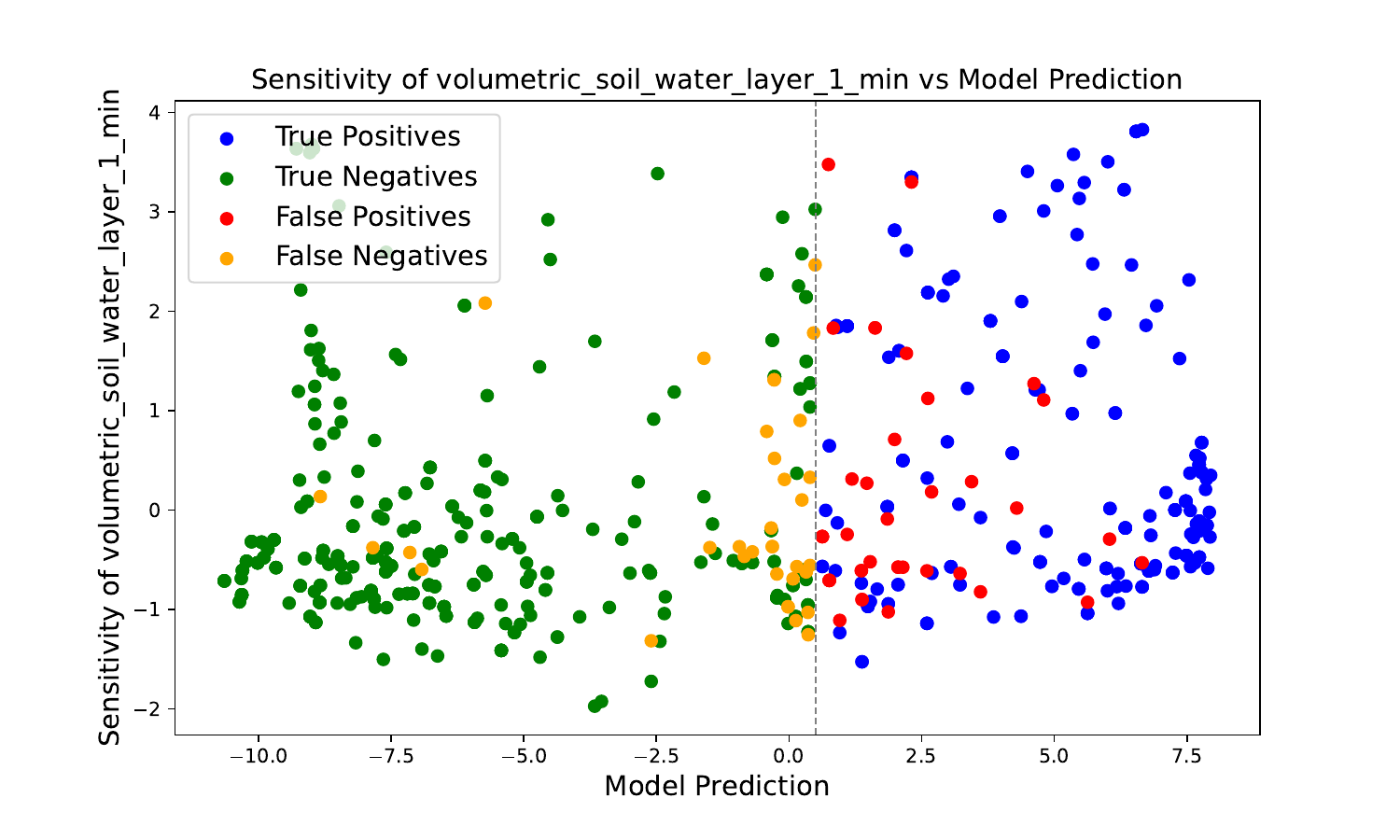}
        \caption{Volumetric Soil Water Layer 1 Min}
        \label{fig:volumetric_soil_water_layer_1_min}
    \end{subfigure}
\end{figure}

\begin{figure}
    \centering
    \includegraphics[width=\linewidth]{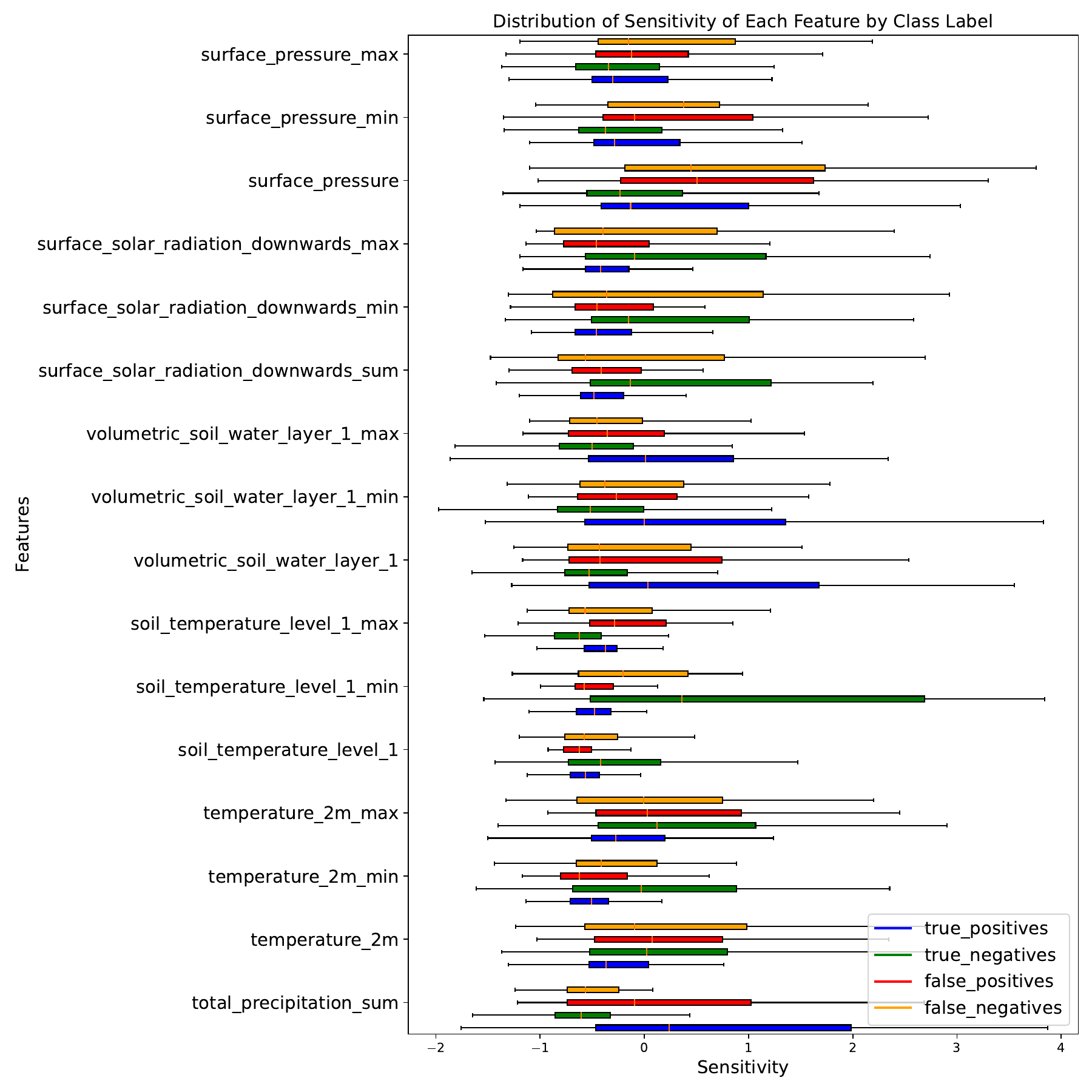}
    \caption{This plot replicates Figure~\ref{fig:sensitivity-horizontal}, but vertically, allowing for ease of reading.}
    \label{fig:sensitivity-vertical}
\end{figure}

\begin{itemize}
    \item In Figure~\ref{fig:total_precipitation_sum}, we look at the Total Precipitation Sum. This is accumulated liquid and frozen water like rain and snow, stemming from both large-scale weather patterns and convective activities, measured in depth (meters) and representing the depth water would have if spread evenly over a grid box. We note the upward trend in sensitivity among the true positive class (in blue) as the model predictions get higher. This suggests that higher likelihood green-ups (i.e. higher model predictions) are especially sensitive to reduced precipitation to the mean. 
    \item In Figures~\ref{fig:temperature_2m},~\ref{fig:temperature_2m_max} and~\ref{fig:temperature_2m_min}, we examine Temperature (2 meters), as well as Max and Min Temperature (2 meters) respectively. This is air temperature calculated 2 meters above the surface of land, sea, or inland waters, modeled considering atmospheric conditions. These are highly correlated features and thus the trends in their sensitivity are similar: true negative samples are the most highly sensitive samples to this class. As temperature is correlated with precipitation and moisture levels, we conjecture that higher temperatures imply more certainty about no green-up. We further note that this feature had the smallest effects in aggregate on model predictions (see Figure~\ref{fig:sensitivity-vertical}).
    \item In Figures~\ref{fig:solar_radiation_downward_sum},~\ref{fig:solar_radiation_downward_max} and~\ref{fig:solar_radiation_downward_min}, we analyze the sensitivity of Solar Radiation Downward Sum, Max and Min respectively. This represents the amount of solar radiation, including both direct and diffuse, reaching the Earth's surface, accounting for reflection by clouds and aerosols and absorption, approximating what would be measured by a pyranometer. Similarly to temperature, this variable is correlated with precipitation (sun implies no cloud cover, which implies less moisture). Based on the sensitivity plot, we conjecture that the model primarily used this variable in determining true negatives.
    \item In Figures~\ref{fig:surface_pressure},~\ref{fig:surface_pressure_max} and~\ref{fig:surface_pressure_min}, we observe sensitivity of Surface Pressure, Max and Min respectively. This pressure variable, measured in Pascals (Pa), represents the atmospheric pressure on the surface of land, sea, and inland water, reflecting the weight of the air column above, and is utilized alongside temperature to determine air density, with mean sea level pressure commonly used for identifying pressure systems over mountainous areas. We were surprised by the sensitivity of the model to this variable. However, it is known that surface pressure is heavily correlated with weather systems, and can independently effect the rate of plant photosynthesis through disruptions to transpiration \cite{takeishi2013effects}. Still, we found it difficult to interpret the effects of this variable, particularly with respect to false positives and false negatives. We worry that it could be confounding, and may remove it in future model iterations.
    \item In Figures~\ref{fig:soil_temperature_level_1},~\ref{fig:soil_temperature_level_1_max} and~\ref{fig:soil_temperature_level_1_min}, we study the sensitivity of Soil Temperature (Level 1), Max and Min respectively. The temperature of soil in layer 1 (0 - 7 cm) is set at the middle of each layer, with heat transfer calculated at the interfaces between them, and an assumption of no heat transfer out of the bottom of the lowest layer. Most samples were not particularly sensitive to this variable, where others were. The sensitivities suggests that the max soil temperature was useful in deciding positive classifications, while the min soil temperature was useful in negative classifications. The soil temperature likely correlates with water volume and solar heat, which partially explains the sensitivities. 
    \item In Figure~\ref{fig:volumetric_soil_water_layer_1},~\ref{fig:volumetric_soil_water_layer_1_max} and~\ref{fig:volumetric_soil_water_layer_1_min}, the Volumetric Soil Water (Layer 1), Max and Min is examined. The volumetric soil water in soil layer 1 (0 - 7 cm) is related to the soil texture, depth, and the underlying groundwater level, with the surface at 0 cm. Soil water is vital for plant growth and thus for accurate positive predictions - an upward trend in model likelihood scores correlated with higher sensitivity is apparent in all three sensitivity plots.
\end{itemize}

\begin{figure}
    \centering
    \includegraphics[width=\linewidth]{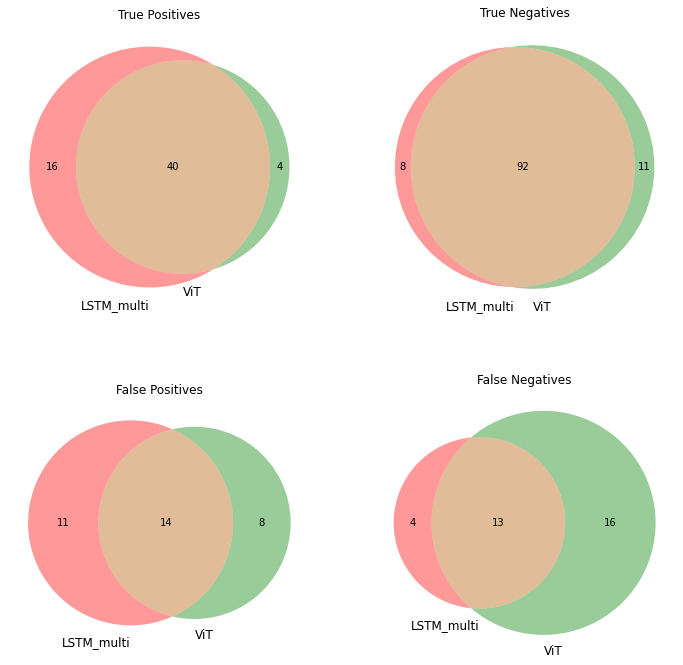}
    \caption{Understanding the overlapping correctness between our \textbf{LSTM (multi)} and \textbf{ViT} models informed our decision to pursue a multi-modal approach. Above, you can see the overlapping correctness for the two models, for each class in the confusion matrix. Though both models agree on at least a plurality of predictions in each case, their predictive correctness does not overlap entirely. In fact, the types of errors each model makes are vastly different, and suggests to us that non-overlapping signal might be captured by using an ensemble or multi-modal approach. Though we did not succeed in creating a multi-modal approach that fully utilized this signal, we believe that future improvements in green-up accuracy will be dependent on combining direct satellite observation with climate and other remote sensing data in this way.}
    \label{fig:venn}
\end{figure}


\begin{figure}[ht]
    \centering

    \begin{subfigure}{.45\textwidth}
        \centering
        \includegraphics[width=\linewidth]{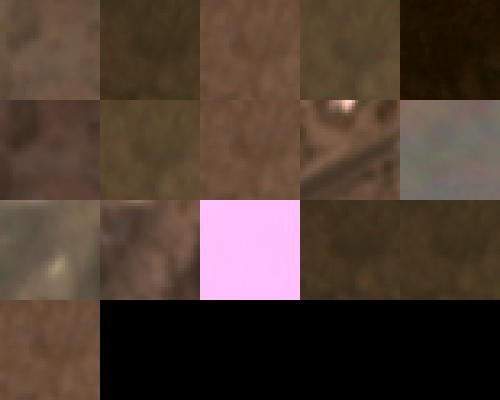}
        \caption{LSTM Unique True Positives (ViT False Negatives).}
    \end{subfigure}
    \quad
    \begin{subfigure}{.45\textwidth}
        \centering
        \includegraphics[width=\linewidth]{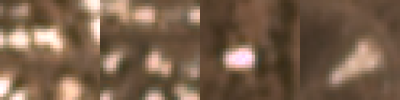}
        \caption{ViT Unique True Positives (LSTM False Negatives).}
    \end{subfigure}
    
    \begin{subfigure}{.45\textwidth}
        \centering
        \includegraphics[width=\linewidth]{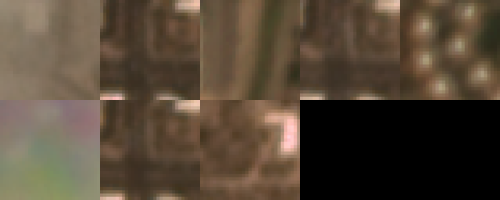}
        \caption{LSTM Unique True Negatives (ViT False Positives).}
    \end{subfigure}
    \quad
    \begin{subfigure}{.45\textwidth}
        \centering
        \includegraphics[width=\linewidth]{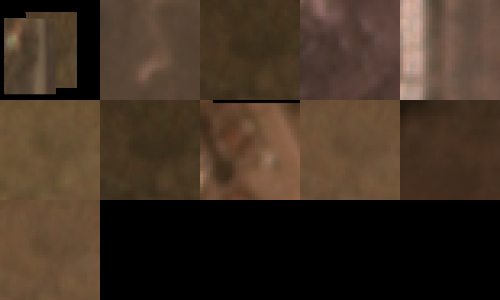}
        \caption{ViT Unique True Negatives (LSTM False Positives).}
    \end{subfigure} \\
\caption{To provide a sense of the types of mistakes the \textbf{ViT} model makes in particular, we provide the following patches of satellite imagery, corresponding to confusion matrix sections of the Venn diagram in Figure~\ref{fig:venn}. You can see that images the \textbf{ViT} model ``missed'' (i.e. false negatives) in relation to the \textbf{LSTM (multi)} model are almost entirely due to poor image quality or cloud cover. Meanwhile, unique \textbf{LSTM} true negatives and unique \textbf{ViT} true positives have similar image quality, suggesting the model is inherently biased towards higher quality images for positive classification. This type of bias would be ameliorated through the collection of more data and better image cleaning procedures, as well as higher resolution imagery.}
\label{fig:overlap_images}
\end{figure}

\end{document}